\documentclass[10pt,twocolumn,letterpaper]{article}
\usepackage[pagenumbers]{cvpr} 

\makeatletter
\def\@LN@col#1{}
\def\@LN#1#2{}
\let\@LN@vline\relax
\makeatother

\usepackage{multirow}
\usepackage{booktabs}
\usepackage{amsmath}
\usepackage{colortbl}
\usepackage{tcolorbox}
\usepackage{ulem}

\definecolor{cvprblue}{rgb}{0.21,0.49,0.74}
\usepackage[pagebackref,breaklinks,colorlinks,allcolors=cvprblue]{hyperref}
\def\paperID{7788} 
\def\confName{CVPR}
\def\confYear{2025}

\title{Self-Supervised Pre-training with Combined Datasets for 3D Perception in Autonomous Driving}

\author{
Shumin Wang \and Zhuoran Yang \and Lidian Wang \and Zhipeng Tang \and Heng Li \and Lehan Pan \and Sha Zhang \and Jie Peng \and Jianmin Ji \and Yanyong Zhang\\
University of Science and Technology of China\\
Hefei, China\\
{\tt\small \{shuminwang, shanpoyang, lidianw, tangzhipeng, li\_heng, lhpan, zhsh1, pengjieb\}@mail.ustc.edu.cn}, \\ 
{\tt\small \{jianmin@ustc.edu.cn, yanyongz\}@ustc.edu.cn}
}

\begin{document}
\maketitle
\begin{abstract}
The significant achievements of pre-trained models leveraging large volumes of data in the field of NLP and 2D vision inspire us to explore the potential of extensive data pre-training for 3D perception in autonomous driving.
Toward this goal, this paper proposes to utilize massive unlabeled data from heterogeneous datasets to pre-train 3D perception models. We introduce a self-supervised pre-training framework that learns effective 3D representations from scratch on unlabeled data, combined with a prompt adapter based domain adaptation strategy to reduce dataset bias. The approach significantly improves model performance on downstream tasks such as 3D object detection, BEV segmentation, 3D object tracking, and occupancy prediction, and shows steady performance increase as the training data volume scales up, demonstrating the potential of continually benefit 3D perception models for autonomous driving. We will release the source code to inspire further investigations in the community.
\end{abstract}    
\section{Introduction}
\label{section:intro}
In recent years, the fields of NLP~\cite{touvron2023llama,bai2023qwen} and 2D vision~\cite{dehghani2023scaling,radford2021learningtransferablevisualmodels,he2019moco} have witnessed tremendous success through pre-training models on a large volume of data. However, in the domain of 3D vision, particularly in the area of 3D perception for autonomous driving, this trend is difficult to achieve due to the lack of properly annotated datasets. Even though many public datasets~\cite{caesar2020nuscenes,mandal2021lyft,sun2020scalability,mao2021one,caesar2022nuplanclosedloopmlbasedplanning} are available, a large portion of the data within them is unlabeled~\cite{mao2021one,caesar2022nuplanclosedloopmlbasedplanning} because annotating such 3D data is highly complex and time-consuming, rendering annotation far slower than the rate at which data is collected. On the other hand, the data exhibit significant distribution differences~\cite{wu2024towards} because 3D datasets are often collected independently in different road environments and with different sensor configurations. This poses a challenge for the collaborative use of multiple datasets to further expand the scale of training data.

In this work, we attempt to address the scarcity of annotated data by developing a self-supervised pre-training approach on a combination of heterogeneous datasets. Specifically, we focus on multi-modal 3D perception models because public datasets today are collected by different sensors. Furthermore, we note that state-of-the-art 3D perception models on most tasks and benchmarks such as BEVFusion~\cite{liu2023bevfusion}, CMT~\cite{Yan_2023_ICCV_cmt}, SparseLIF~\cite{zhang2024sparselif} and UniTR~\cite{wang2023unitr} are all multimodal models that combine camera images and LiDAR point clouds. Thus, our work focuses on LiDAR-camera fusion-based 3D perception models.

Firstly, we extract large amounts of unlabeled data from publicly available datasets including NuScenes~\cite{caesar2020nuscenes}, Lyft~\cite{mandal2021lyft}, and ONCE~\cite{mao2021one} to pre-train models, focusing on investigating the impact of large-scale unlabeled pre-training data on 3D perception models. To achieve this, we require a self-supervised training method that enables the backbone networks of both image and point cloud modalities to co-evolve from scratch on vast amounts of data.
Although previous researches have proposed numerous self-supervised pre-training approaches for images~\cite{caron2021emerging,he2021maskedautoencodersscalablevision,wei2022masked,xie2022simmim}, point clouds~\cite{pointcontrast,zhang2022point,hess2023masked,pang2022masked} and aligning the two modalities~\cite{sautier2022imagetolidarselfsuperviseddistillationautonomous,zhou2023unidistilluniversalcrossmodalityknowledge,liu2023geomim,sun2023calico,li2022simipu}, they often focus either on single modalities~\cite{caron2021emerging,he2021maskedautoencodersscalablevision,wei2022masked,xie2022simmim,pointcontrast,zhang2022point,hess2023masked,pang2022masked}, cross-modal distillation~\cite{chen2022bevdistill,zhou2023unidistilluniversalcrossmodalityknowledge,liu2023geomim}, or leveraging pre-trained networks~\cite{chen2022bevdistill}, which does not fully satisfy our requirements.

We therefore propose a self-supervised pre-training framework for collaborative learning between point cloud and image data. In this framework, both modalities are unified in the Bird's Eye View (BEV) perspective, where contrastive learning facilitates mutual enrichment of information across modalities, facilitating collaborative knowledge acquisition from large-scale data. To mitigate potential semantic information loss when elevating images to BEV, we introduce a Masked Autoencoder (MAE) loss for the image modality, ensuring precise semantic capture. Our approach is designed to train backbones of both image and point cloud modalities from scratch, making it well-suited for investigating the effects of large-scale data pre-training on multi-modal 3D perception models.

Secondly, we address the domain gap between datasets. 3D data for autonomous driving are typically collected independently using specialized equipment in different road environments, resulting in varied datasets. These datasets may come from different public sources or be privately collected by companies and organizations at various locations, times, and using different equipment. As a result, data from different sources may exhibit significant distributional differences. The distributional bias may lead to training conflicts, resulting in suboptimal outcomes when they are combined for model training. This hinders the further expansion of the training data scale for 3D perception models.

To mitigate the impact of domain gap between datasets, we employ a prompt adapter-based training strategy~\cite{wu2024towards} to disentangle dataset biases from the backbone network. We set learnable prompt parameters for different datasets, which are activated separately during the training process and connected to the backbone network via an adapter. This method allows the model to leverage larger and more diverse data sources more effectively.

In this paper, we propose a self-supervised pre-training framework which can leverage heterogeneous unlabeled datasets for 3D perception models in autonomous driving. Our framework is designed to reduce dependency on large annotated datasets and enhance the data scalability in 3D perception domain. The results in four downstream tasks, 3D object detection, 3D object tracking, BEV segmentation, and occupancy prediction demonstrate that our method can effectively improve model performance with heterogeneous unlabeled datasets. Meanwhile, through our pre-training method, the performance of 3D perception models shows a steady improvement as the amount of training data increases, demonstrating the potential to expand the frontier of 3D perception models.

To summarize, we make the following contributions: 
\begin{itemize}
    \item We propose a self-supervised pre-training method which is able to learn effective 3D representations from scratch on a combination of heterogeneous datasets.
    
    \item We evaluate our model on four tasks: 3D object detection, 3D object tracking, BEV segmentation and occupancy prediction, and demonstrate the effectiveness of our method.

    \item We scale up the training data volume to 250,000 frames. We observe that the models demonstrate steady performance improvement as the training data volume scales up, which suggests that our method can be further scaled with additional training resources.
\end{itemize}

\section{Preliminary}
\label{section:preliminary}
In this section, we present 
the idea of self-supervised pre-training and categorize the proposed approaches for 3D perception into two groups.
\paragraph{Self-supervised pre-training}
Self-supervised pre-training is a method in which models are trained by constructing pretext tasks, with training labels derived from the data itself. This allows models to learn preliminary knowledge from unlabeled data and converge more efficiently on downstream tasks. In the field of 3D perception, this method is commonly employed to train backbone encoders~\cite{zhang2024hvdistill}, enabling them to acquire accurate and effective feature encoding capabilities, which can be utilized in subsequent downstream tasks.

In recent studies on 3D perception 
, the pretext tasks for self-supervised pre-training are typically divided into two categories: contrastive learning and masked autoencoders. We will introduce both approaches in detail below.
\paragraph{Contrastive learning}
Contrastive learning aims to learn representations by contrasting positive and negative data pairs. This can be applied to representation learning within a single modality~\cite{he2019moco,pointcontrast} or to align different modalities~\cite{radford2021learningtransferablevisualmodels,sun2023calico}. When applied to a single modality, different views of the same data frame are typically treated as positive samples, while different frames serve as negative samples. This approach helps the model to learn semantic information by supervising it to accurately differentiate data frames. In the case of aligning different modalities, matching pairs of data from the dataset (e.g., point clouds and images at the same moment) are considered positive samples, while non-matching pairs are treated as negative samples, allowing the model to learn similar semantic representations across modalities. 
\paragraph{Masked-autoencoder}
The Masked-Autoencoder (MAE) method~\cite{he2021maskedautoencodersscalablevision} involves masking a portion of the input image or point cloud, after which the encoder model encodes the remaining visible parts into a feature sequence. An additional decoder module is then used to reconstruct the masked portion. The MAE approach helps the model learn semantic information and local features embedded in the image or point cloud by making it predict the masked regions based on the unmasked parts. This enables the network to extract meaningful features, improving perception accuracy on downstream tasks.

\section{Method}
\label{section:method}

\subsection{Design of training losses}
\label{section:pretrain_overall}

To effectively utilize unlabeled data, we design two kinds of losses in our framework. 
Firstly, we perform contrastive learning on the BEV features of both image and point cloud modalities. This process allows the unique features within each modality to mutually inspire and guide the model toward learning more effective feature representations.
Secondly, as images are collected as a 2D space, there is a risk of losing semantic information during the process of elevating it to 3D and transforming it into BEV. We then apply a self-supervised loss to maintain their semantic information. Previous research~\cite{park2023self} has pointed out that MAE, compared to contrastive learning-based methods performs better in scenarios involving complex decoder fine-tuning, which aligns with our use case. We therefore use MAE loss as the self-supervised loss within image modality.

\begin{figure*}[t]
\centering
\includegraphics[width=0.99\linewidth]{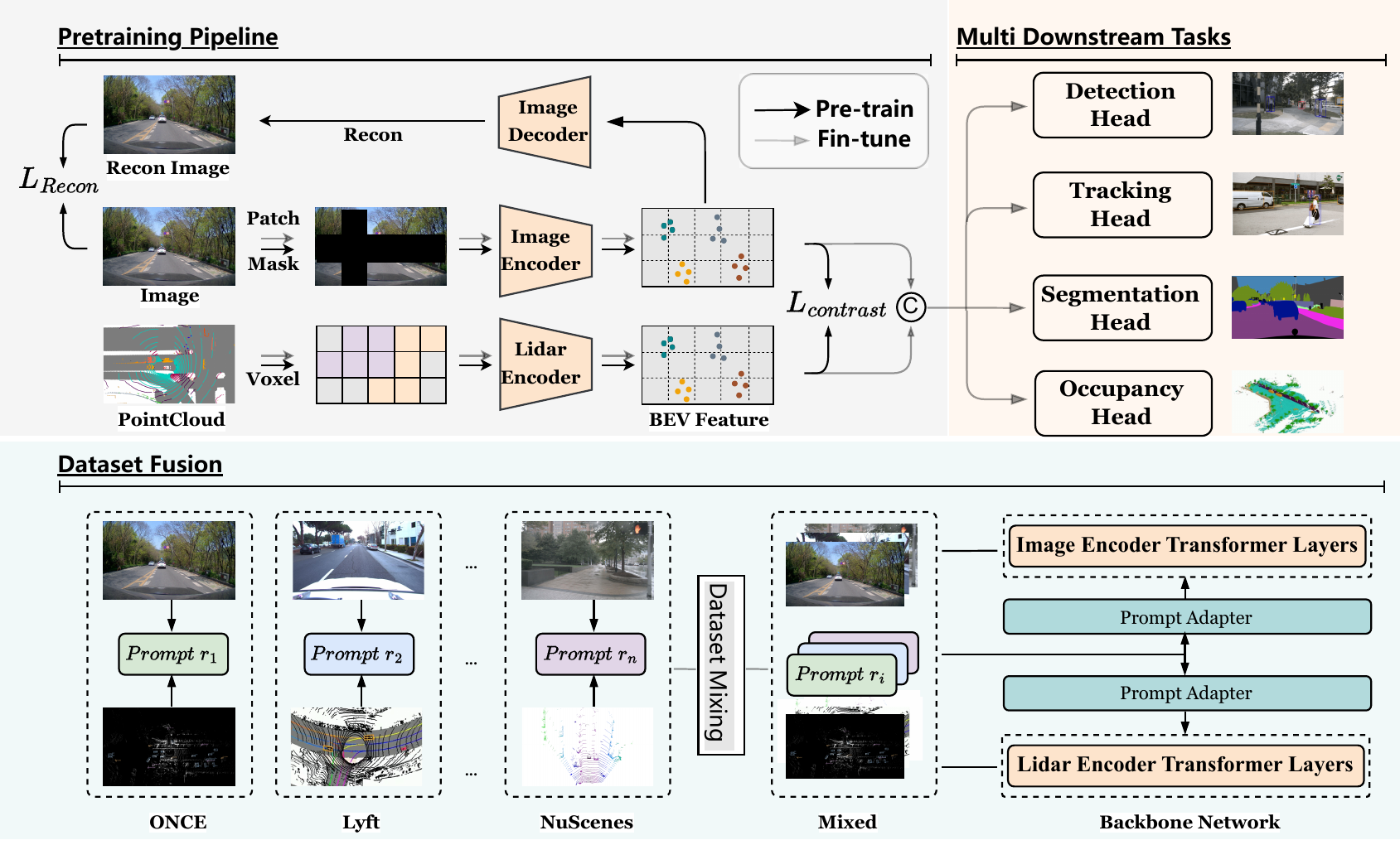} 
\caption{The pre-train-then-fine-tune framework for multi-modal 3D perception integrating image and point cloud data. The point clouds and partially masked images are encoded into BEV feature tokens. We use contrastive loss between the BEV map of the two modalities to make them co-evolve, while use mae loss of recovering the masked portion of images to help capturing their semantic features.}
\label{fig:pretrain}
\end{figure*}

As shown in figure~\ref{fig:pretrain}, our overall training loss is the combination of the two losses, which can be expressed as:
\begin{equation}
\begin{split}
\mathcal{L}_{\text{All}}(\theta;D) = & \mathcal{L}_{\text{MAE}}(\theta_{\text{img}};D_{\text{img}}) \\
+ & \mathcal{L}_{\text{CL}}(\theta_{\text{img}},\theta_{\text{pcd}};D_{\text{img}},D_{\text{pcd}})
\end{split} 
\end{equation}
where $\theta$ represents the model's parameters and $D$ refers to the model's inputs. $D_{img}$ and $D_{pcd}$ refers to input data of images and point clouds, respectively. While $\theta_{img}$ and $\theta_{pcd}$ refers to the parameters of sub-network which process $D_{img}$ and $D_{pcd}$. We then introduce the two training losses within our framework in detail:

\paragraph{Contrastive loss} 
Each sample \( D_i \) from dataset \( \mathcal{D} \) consists of paired image and point cloud data, \( D_i = \{ I_i, P_i \} \). After serialization, the data from both modalities are fed into their respective backbone networks, \( f_{\text{img}} \) for the image and \( f_{\text{pcd}} \) for the point cloud. The point cloud features can be directly transformed into the BEV (Bird's Eye View) perspective, while the image features are projected into the BEV space using the LSS (Lift, Splat, Shoot) network, leveraging the camera extrinsic parameters \( \theta_{ce} \), i.e.
\begin{equation}
    \begin{split}
        BEV_{img} &= LSS(f_{img} (I_i), \theta_{ce})\\
        BEV_{pcd} &= f_{pcd} (P_i)
    \end{split}
\end{equation}

The BEV maps are formed with 128 $\times$ 128 grids and we take the grids in corresponding positions \(m \in [0,128\times128)\) in image BEV maps \(BEV_{img}=\{x_i^j\} \)  and point cloud BEV maps \(BEV_{pcd}=\{y_i^j\} \) as positive pairs \((x_i^m,y_i^m)\). Grids in positions not corresponding to each other are treated as negative pairs \((x_i^m,y_i^n),n\neq m\). In the actual training process, due to memory constraints, we randomly sampled some locations from the BEV map for training. We then use NCE loss to train the model and the loss is formulated as:
\begin{equation}
    \mathcal{L}_{\text{CL}} = -\sum_{i=1}^{|\mathcal{D}|}\frac{1}{|\mathcal{D}|}\sum_{j=1}^{K}\frac{1}{K}\log \frac{\exp(x_i^j \cdot y_i^j / \tau)}{\sum_{k=1}^{K} \exp(x_i^j \cdot y_i^k /\tau )}
\end{equation}
where $ \tau $ is temperature parameter, and $ K $ is the number of grids sampled from BEV maps for contrastive learning.

\paragraph{MAE loss}
Each of the input images \(I_i\) from dataset \(\mathcal{D}\) is divided into patches \(c_i = \{c_i^j\}\), and we randomly sample a portion of the patches to be masked (i.e. set to zero). After that, all patches including the masked and unmasked ones will be passed through the SwinTransformer encoder to embed them into image tokens. After that, an additional MAE decoder processes these tokens, reconstructing the original images. The reconstruction quality is assessed using the MAE loss, guiding the network to learn meaningful image features in a self-supervised manner:

\begin{equation}
    \mathcal{L}_{\text{MAE}} = \frac{1}{|\mathcal{D}|} \sum_{i=1}^{|\mathcal{D}|} \| \hat{c}_i - c_i \|_2^2,
\end{equation}
where $ \hat{c}_i $ is the reconstructed image patches, and $ c_i $ is the original image patches.

\subsection{Dataset prompt formulation}
\label{section:prompt_form}

\begin{figure}[t]
\centering
\includegraphics[width=1.1\columnwidth]{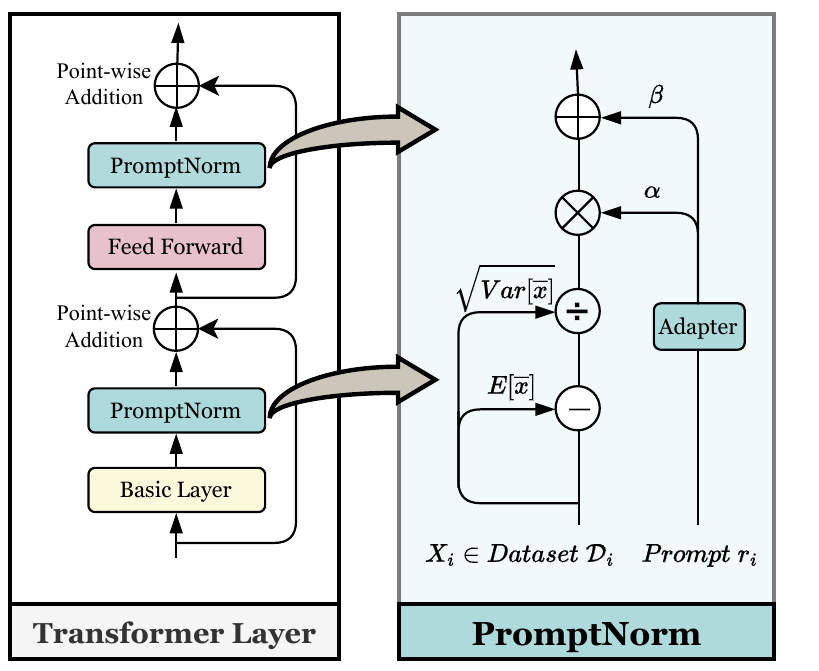} 
\caption{Multi-dataset training strategy with prompt adapters. We set tunable prompts for each dataset and mix the dataset during training. The prompts are injected into the backbones with MLP adapters.}
\label{fig:fusion}
\end{figure}

As discussed in section \ref{section:intro}, naively mixing multiple 3D perception datasets for model training may lead to suboptimal performance because of the domain gap between different datasets. Therefore, we employed a Prompt Adapter design, inserting additional learnable parameters related to specific datasets into the backbone networks of both image and point cloud modalities to retain the common knowledge of different datasets within the backbone model while separating the differential information between datasets outside it. (shown in figure \ref{fig:fusion}). Specifically, during multi-dataset training, we evenly mix data from different datasets and reindex them. For each dataset $\mathcal{D}_i$, we define a set of learnable parameters $r_i$ as corresponding soft prompt. During training, each data sample activates the corresponding prompt parameters for computation based on its index.

Following~\cite{wu2024towards}, we employ an MLP adapter network to map the prompt into weight and bias parameters. These parameters are then integrated with the layer\_norm of the transformer layers in the backbone network through linear operations, thus injecting the prompt into the backbones.

Therefore, for each input sample $d_i^j \in \mathcal{D}_i$, the normalization process of the data sample in the backbones can be expressed as:
\begin{equation}
\begin{aligned}
      \alpha,\beta &= Adapter(r_i),\\
    PromptNorm(d_i^j) &= \alpha \times LayerNorm(d_i^j) + \beta
\end{aligned}
\end{equation}

\subsection{Optimization objective}
\label{section:prompt_object}
We replace all LayerNorm layers in both the image and point cloud backbone networks with PromptNorm. Therefore, the trainable parameters during the training process include the backbone network parameters \(\theta_b\), the soft prompt \(C = \{C_i\}\), and the adapter parameters \(\theta_a\).

Therefore, the final training objective becomes:
\begin{equation}
 argmin_{ r,\theta_a,\theta_b}\frac{1}{N}\sum\limits_{i=1}^{N}\frac{1}{|\mathcal{D}_i|}\sum\limits_{d_i^j \in \mathcal{D}_i}\mathcal{L}_{All}(d_i^j;r_i;\theta_a,\theta_b)   
\end{equation}
where N is the number of datasets, \(\theta_a\) is the parameters of adapters and \(\theta_b\) is the parameters of backbones.

In such optimization process, the soft prompt \(r\) can be optimized for efficient representation of dataset-specific information, while the backbone network learns general knowledge and feature representations across datasets. The adapter parameters are optimized for the effective connection between the two for specific datasets.

\section{Experiment \& results}
\label{section:exp}
In this section, we first introduce the settings of our experiments. Then, we show our evaluation results of our models to demonstrate how our method leverages large-scale unlabeled data for pre-training to continually improve the performance of autonomous driving 3D perception models. We will answer following research questions (RQs):

\textbf{RQ1}: Can our proposed self-supervised pre-training framework effectively improve model performance on downstream tasks? (Ref. Section~\ref{section:exp_pretrain_effect})

\textbf{RQ2}: Can the prompt adapter technique we employ help mitigating the domain gap between 3D datasets in our self-supervised pre-training framework? (Ref. Section~\ref{section:exp_dataset_fusion})

\textbf{RQ3}: Does our proposed approach demonstrate the potential for continuous improvement in performance as the scale of data increases? (Ref. Section~\ref{section:exp_scaling})

\subsection{Experiment settings}
\paragraph{Models}
\label{subsection:exp_model}
In recent 3D perception research~\cite{liu2023bevfusion,li2022bevformer,mao2021voxel,fan2022embracing,wang2023dsvtdynamicsparsevoxel}, transformer networks are gradually gaining popularity due to their stronger scalability and better performance. We adopt the Swin Transformer~\cite{liu2021swintransformerhierarchicalvision} and DSVT~\cite{wang2023dsvtdynamicsparsevoxel} architectures for the backbones of image and point cloud modalities, respectively.
Swin transformer is one of the most popular transformer backbones in computer vision while DSVT is one of the state-of-the-art backbones for point cloud and is also widely used in 3D perception~\cite{wang2023unitr,yang2024pred,yan2023spot}. For encoding image features into BEV, we employ the LSS method used in BEVFusion~\cite{liang2022bevfusion}. 
\paragraph{Datasets}
During the pre-training stage, we use the Nuscenes~\cite{caesar2020nuscenes}, Lyft~\cite{mandal2021lyft}, and ONCE~\cite{mao2021one} datasets. All of them are widely used large-scale 3D perception datasets, offering 360-degree surround-view image data and point cloud data collected by LiDARs. However, there are significant differences in their sensors: the beams for LiDAR varies, being 32, 64, and 40, respectively, and the mounting angles of the cameras also differ.

NuScenes, Lyft and ONCE, datasets provide 28,000, 22,000 and 980,000 training frames, respectively. 
To control the scale of the experiments, we sampled 50,000 frames from the ONCE dataset when investigating the multi-dataset training strategy. 
In the dataset scaling experiments, we sampled 200,000 frames from ONCE and balanced the number of training samples across datasets by repeating the data from NuScenes and Lyft four times. To align the data cross different datasets, we standardized the 6-camera surround-view visual signals and unified the LiDAR range and feature dimensions, but did not adjust the density of the points to maintain as much raw information as possible.
Detailed configurations are provided in Appendix A.

For downstream benchmark, we propose two kinds of settings. In the main setting, we use 20\% of the NuScenes dataset to evaluate the model’s capabilities with a limited number of samples with annotation. This behchmark includes four downstream tasks: 3D object detection, 3D object tracking, BEV segmentation, and occupancy prediction. We also provide a held-out setting in which we use 20\% protion of the Waymo Open~\cite{sun2020scalability} dataset as out-of-domain experimental data to validate whether our method works when the fine-tuning data distribution differs from the pre-training data. This bechmark only officially supports the 3D object detection task. Model structure and training detailes are provided in Appendix C. 
\paragraph{Evaluation}
In 3D object detection task, we utilize official evaluation metric of NuScenes, Mean Average Precision (mAP) and nuScenes Detection Score (NDS), respectively reflecting accuracy rating and comprehensive consideration of translation, scale, angle, and velocity. For Waymo Open dataset, we utilize L1 mAP and L1 mAPH, which are also officially provided. 
In the 3D object tracking task, we also use the official evaluation metrics of NuScenes, Average Multi Object Tracking Accuracy (AMOTA) and Average Multi Object Tracking Precision (AMOTP), which average the MOTA and MOTP metrics at different recall thresholds. MOTA measures the accumulation of tracking errors, including false positives, missed targets, and identity switches while MOTP evaluates the misalignment between annotated and predicted bounding boxes.
For BEV segmentation and occupancy prediction, we adopt mean Intersection-over-Union (mIoU) which is computed as the ratio of the intersection to the union of prediction and ground truth and averaged between classes. An overall IoU is also provided fro occupancy prediction task.

\subsection{Effectiveness of pre-training framework}
\label{section:exp_pretrain_effect}
\begin{table*}
    \centering
    \caption{Evaluation of our pre-training framework with baseline methods on downstream tasks. The models are pre-trained on NuScenes dataset and fine-tuned on 20\% of NuScenes dataset. To fit the scope of the paper, we only use the self-supervised loss when reproducing BEVDistill. Hyperparameters are provided in Appendix A.}
    \resizebox{\textwidth}{!}{
    \begin{tabular}{lccccccc}
    \toprule
        \multirow{2}*{Pre-training Method}& \multicolumn{2}{c}{3D Object Detection}& \multicolumn{2}{c}{3D Object Tracking} & BEV Segmentation &\multicolumn{2}{c}{Occupancy Prediction}   \\
        \cmidrule(r){2-3}\cmidrule(r){4-5}\cmidrule(r){6-6}\cmidrule(r){7-8}
        
        & mAP$\uparrow$&NDS$\uparrow$&AMOTA$\uparrow$&AMOTP$\downarrow$ &mIoU$\uparrow$&mIoU$\uparrow$&IoU$\uparrow$\\
        \midrule
        Scratch & $51.5_{\pm 0.8}$ & $53.5_{\pm 0.5}$ & $57.1_{\pm 1.1}$ & $80.9_{\pm 0.9}$ & $39.1_{\pm 0.3}$ & $15.7_{\pm 0.4}$ & $32.3_{\pm 0.6}$\\
        BEVDistill &$54.1_{\pm 0.1}$ &$57.5_{\pm 0.6}$ & $60.6_{\pm0.0}$&$76.5_{\pm 0.6}$ & $39.1_{\pm 0.1}$ &$15.8_{\pm 0.1}$ &$32.3_{\pm 0.3}$ \\
        \rowcolor{gray!20}
        Ours & $\textbf{55.5}_{\pm 0.5}$ & $\textbf{58.8}_{\pm 0.5}$ & $\textbf{61.8}_{\pm 0.7}$ & $\textbf{73.2}_{\pm 1.2}$ & $\textbf{39.3}_{\pm 0.1}$ &$\textbf{15.9}_{\pm 0.2}$ &$\textbf{32.5}_{\pm 0.1}$ \\
        \bottomrule
   
    \end{tabular}}
    
    \label{tab:pretrain_effect}
\end{table*}

We first evaluate the effectiveness of our pre-training framework when conducted on a single dataset. We use NuScenes dataset alone to pre-train our model and compare it with following baselines:
\begin{itemize}
    \item Scratch: Randomly initialize the networks without any pre-training.
    \item BEVDistill: State-of-the-art open-source pre-training method similar to our setting. Though it was designed to distill knowledge from pre-trained image networks, it can also be adopted to train models from scratch~\cite{chen2022bevdistill,sun2023calico}.
\end{itemize}
More discussions about baselines and related works are provided in Appendix B.
\paragraph{Main results \& ablation study}Table \ref{tab:pretrain_effect} presents a comparison of the performance on various downstream tasks between our method and baseline methods in the held-in setting. In all four tasks, our pre-training framework achieves significant performance improvements and outperforms the baseline methods. To address whether both the losses in our framework contributes to the performance gain, we provide ablation study by removing one of the loss functions. As shown in Table \ref{tab:pretrain_ablation_loss}, both loss functions contribute to the improvement of models' performance on downstream tasks.
\begin{table}
    \centering
    \caption{Ablation study about the contributions of the losses we use in our pre-training framework. The models are evaluated on 3D object detection and BEV segmentation task.}
    \begin{tabular}{lccc}
    \toprule
        \multirow{2}*{Loss func.}& \multicolumn{2}{c}{3D object det.}& BEV seg.\\
        \cmidrule(r){2-3}\cmidrule(r){4-4}
         & mAP &NDS &mIoU  \\
         \midrule
         Full & $\textbf{55.5}_{\pm 0.5}$ & $\textbf{58.8}_{\pm 0.5}$& $\textbf{39.3}_{\pm 0.1}$ \\ 
         -$\mathcal{L}_{MAE}$ &$54.1_{\pm 0.9}$ & $58.0_{\pm 0.8}$ & $39.1_{\pm 0.3}$ \\ 
         -$\mathcal{L}_{CL}$ & $53.0_{\pm 0.4}$ & $57.2_{\pm 0.3}$ & $39.2_{\pm 0.2}$ \\ 
         \bottomrule
    \end{tabular}
    
    \label{tab:pretrain_ablation_loss}
\end{table}
\paragraph{Results in held-out setting}
We fine-tune the NuScenes-pre-trained models on the Waymo dataset for 3D object detection task and test their performance. This will show the robustness of different methods against domain shift. Results in table \ref{tab:pretrain_ood} show that our method also effectively enhances model performance in out-of-distribution scenarios and outperforms baseline methods, demonstrating strong generalization capabilities.
\begin{table}
    \centering
    \caption{Evaluation on 3D object detection task about the robustness of our pre-training framework when the domain shifts. The models are pre-trained on NuScenes dataset and fine-tuned on 20\% of Waymo dataset.}
    \begin{tabular}{lcc}
    \toprule
         Pre-training Method& L1 mAP &L1 mAPH  \\
         \midrule 
         Scratch&$65.9_{\pm 1.1}$&$61.9_{\pm 0.9}$\\ 
         BEVDistill&$64.8_{\pm 1.1}$ &$60.9_{\pm 1.1}$ \\ 
         Ours&$\textbf{66.6}_{\pm 0.8}$&$\textbf{62.6}_{\pm 0.7}$\\ 
         \bottomrule
    \end{tabular}
    \label{tab:pretrain_ood}
\end{table}

\begin{tcolorbox}[colback=gray!10, colframe=black, title=RQ1 Summary:]
Our proposed pre-training framework can effectively improve models' performance on various downstream tasks with the contribution of all losses in both in-domain and out-of-domain scenarios.
\end{tcolorbox}

\subsection{Multi dataset pre-training}
\label{section:exp_dataset_fusion}
In this section, we show that the domain gap between datasets would cause suboptimal model performance and then investigate how our multi-dataset training strategy successfully address this issue. To achieve this, we conduct pre-training under the following configurations: 
\begin{itemize}
    \item  Single dataset: Pre-training using only NuScenes dataset.
    \item  Multi-dataset w/o prompt: Pre-training on a naive combination of NuScenes, Lyft, and ONCE datasets.
    \item  Multi-dataset w/ prompt: Pre-training with NuScenes, Lyft, and ONCE, utilizing our prompt training strategy.
\end{itemize}

\paragraph{Main results}Based on the results in Table \ref{tab:dataset_fusion}, simply mixing the datasets for pre-training do not effectively improve the model's performance on downstream tasks, and even perform worse on some tasks than using a single dataset. However, using the prompt-based multi-dataset training strategy lead to a significant performance improvement in all downstream tasks, proving it's effectiveness of mitigating the domain gap between 3D datasets.
\begin{table*}
    \centering
    \caption{Evaluation on downstream tasks with different pre-training strategies conducted. NuScenes only represents the model is pre-trained using only NuScenes dataset. The others are pre-trained on the fusion dataset combined with NuScenes, Lyft and ONCE with or without dataset prompt. The evaluation is conducted on NuScenes dataset. Hyperparameters are provided in Appendix A.}
    \begin{tabular}{lccccccc}
    \toprule
        \multirow{2}*{Pre-training Method}& \multicolumn{2}{c}{3D Object Detection}& \multicolumn{2}{c}{3D Object Tracking} & BEV Seg. &\multicolumn{2}{c}{Occupancy Prediction}\\
        \cmidrule(r){2-3}\cmidrule(r){4-5}\cmidrule(r){6-6}\cmidrule(r){7-8}
        
        & mAP$\uparrow$&NDS$\uparrow$&AMOTA$\uparrow$&AMOTP$\downarrow$ &mIoU$\uparrow$&mIoU$\uparrow$&IoU$\uparrow$\\
        \midrule
        Single dataset&$\uline{53.3}_{\pm 0.4}$&$54.2_{\pm 2.0}$&$\uline{58.7}_{\pm 0.4}$&$\textbf{77.2}_{\pm 1.5}$&$38.9_{\pm 0.1}$&$\uline{16.0}_{\pm 0.4}$&$\uline{32.4}_{\pm 0.3}$\\
        Multi-dataset w/o prompt&$52.4_{\pm 0.9}$&$\uline{54.5}_{\pm 0.6}$&$58.0_{\pm1.5}$&$78.3_{\pm 1.1}$&$\uline{39.6}_{\pm 0.4}$&$15.9_{\pm 0.1}$&$31.8_{\pm 0.2}$\\
        \rowcolor{gray!20}
        Multi-dataset w/ prompt&$\textbf{53.6}_{\pm 0.4}$ & $\textbf{56.0}_{\pm 0.5}$&$\textbf{59.1}_{\pm 0.5}$&$\uline{78.2}_{\pm 1.2}$&$\textbf{41.4}_{\pm 0.1}$&$\textbf{16.1}_{\pm 0.2}$&$\textbf{32.5}_{\pm 0.1}$\\
        \bottomrule
   
    \end{tabular}
    \label{tab:dataset_fusion}
\end{table*}

\paragraph{Further analysis}One problem people may concern is that during the pre-training phase, we set different tunable prompts for various datasets. However, when applying the pre-trained model to downstream tasks, we do not always use data similar to what was used during pre-training. Thus, the impact of different strategies for using prompts during fine-tuning is crucial. Specifically, when we cannot select an appropriate prompt from those obtained during the pre-training process, is our dataset fusion method still effective?
To address this issue, we conduct experiments with the following strategies during fine-tuning on NuScenes dataset:
\begin{itemize}
    \item W/o prompt: Pre-training with combined dataset without prompt training strategy and then fine-tune.
    \item Correspond: Initialize the prompt of NuScenes with the correct one obtained from pre-training before fine-tuning.
    \item Wrong: Initialize the prompt of NuScenes with the prompt of another dataset(i.e., ONCE) obtained from pre-training before fine-tuning.
\end{itemize}
Results in table \ref{tab:effect_prompt} indicates that even using a mismatched prompt to initialize the model still outperforms a naive dataset mixture. This suggests that our method effectively mitigates conflicts between datasets within the backbone network, preserving generalizable knowledge across datasets.
\begin{table}
    \centering
    \caption{Comparison of different strategies of using prompt during fine-tuning. The evaluations are conducted using NuScenes dataset on 3D object detection and BEV segmentation task.}
    \begin{tabular}{l|ccc}
    \toprule
        \multirow{2}*{Prompt strategie}& \multicolumn{2}{c}{3D object det.}& BEV seg.\\
        \cmidrule(r){2-3}\cmidrule(r){4-4}        
        & mAP&NDS&mIoU\\
        \midrule
        W/o prompt & $52.4_{\pm 0.9}$ & $54.5_{\pm 0.6}$ & $39.6_{\pm 0.4}$\\
        Correspond & $\textbf{53.6}_{\pm 0.4}$ & $\textbf{56.0}_{\pm 0.5}$ & $\textbf{41.4}_{\pm 0.1}$ \\
        Wrong &$\uline{53.1}_{\pm 0.2}$ &$\uline{55.2}_{\pm 0.6}$ & $\uline{41.2}_{\pm 0.1}$ \\
        \bottomrule   
    \end{tabular}    
    \label{tab:effect_prompt}
\end{table}

\begin{figure}
    \centering
    \includegraphics[width=0.99\linewidth]{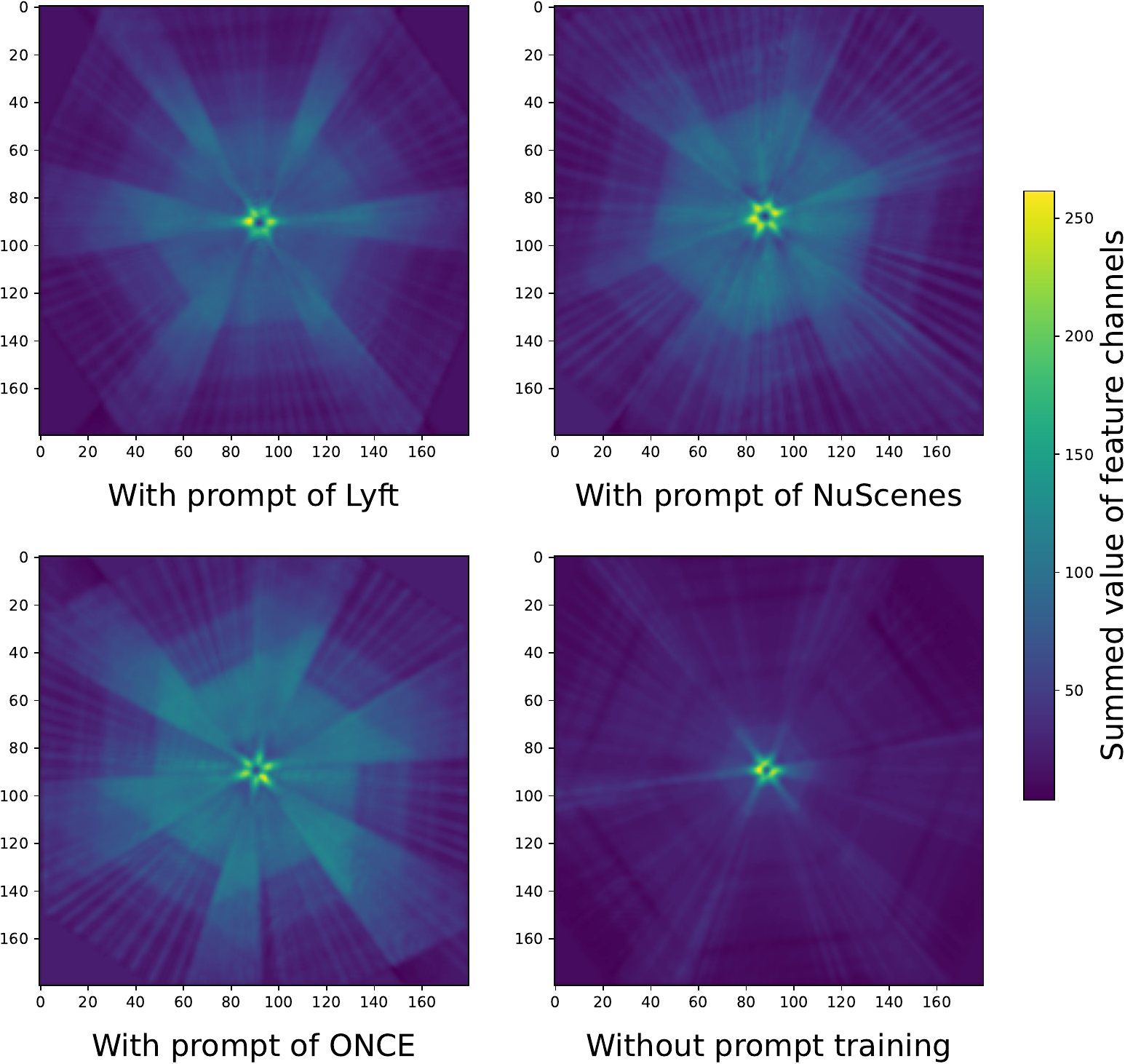}
    \caption{BEV heatmaps of image modality when testing the models by applying various prompts or using the model without prompt training. The x-axis represents the forward and backward direction of the vehicle. The data frame is from Lyft dataset.}
    \label{fig:heatmap}
\end{figure}

To further investigate the impact of prompts, we evaluate the models on a sample from the Lyft dataset, applying various prompts or using the model without prompt training, and visualize the BEV heatmap from the image modality. As shown in Figure~\ref{fig:heatmap}, the prompts enable the model to focus on different angles in space that correspond to the positions of cameras in the dataset. With the correct prompt (i.e., Lyft), the model accurately centers its attention on each camera’s focal area. When there is a mismatch between the data and prompt, the features appear slightly disordered, but can still concentrate on meaningful areas. In contrast, without prompt training strategy, the model struggles to find meaningful feature angles due to conflicts between datasets. Thus, the prompt training strategy effectively mitigates the dataset gap arising from camera angle variations. We provide more experimental results and further analysis in Appendix D.
\begin{tcolorbox}[colback=gray!10, colframe=black, title=RQ2 Summary:]
The prompt adapter technique we employ can effectively help mitigate the domain gap between 3D datasets in our self-supervised pre-training framework.
\end{tcolorbox}
\subsection{Scalability of the pre-training methods}
\label{section:exp_scaling}
The last question we would like to talk about is whether our pre-training method can continually benefit models with the scaling up of data volume. We investigate this issue by conducting experiments on several different training data volumes and compare the model performance on downstream tasks. We progressively expanded the training data volume from 40,000 and 100,000 to 250,000. For the largest experiment, we sample 200,000 frames from ONCE and apply repeated sampling for four times on the NuScenes and Lyft datasets to balance the scale differences between datasets (Details are provided in Appendix A). 

Experimental results in table \ref{tab:dataset_scaling} show that as the training data volume increases, model performance improves steadily. The results indicates that our training method is able to continually improve the performance of 3D perception models with massive training data, thus demonstrating strong scalability.
\begin{table*}
    \centering
    \caption{Evaluation of models' performance on downstream tasks with different scale of pre-training datasets. All experiments are conducted using NuScenes, Lyft and ONCE datasets. Sampling ratios are provided in Appendix A.}
    \begin{tabular}{lccccccc}
    \toprule
        \multirow{2}*{Data Volume}& \multicolumn{2}{c}{3D Object Detection}& \multicolumn{2}{c}{3D Object Tracking} & BEV Segmentation &\multicolumn{2}{c}{Occupancy Prediction}   \\
        \cmidrule(r){2-3}\cmidrule(r){4-5}\cmidrule(r){6-6}\cmidrule(r){7-8}
        
        & mAP$\uparrow$&NDS$\uparrow$&AMOTA$\uparrow$&AMOTP$\downarrow$ &mIoU$\uparrow$&mIoU$\uparrow$&IoU$\uparrow$\\
        \midrule
        40K Frame & $\uline{53.6}_{\pm 0.7}$ & $55.8_{\pm 1.8}$ & $59.0_{\pm 0.2}$ & $\textbf{77.1}_{\pm0.4} $ & $38.9_{\pm 0.2}$ & $15.9_{\pm 0.2}$ &  $32.3_{\pm 0.4}$ \\
        100K Frame & $\uline{53.6}_{\pm 0.4}$ & $\uline{56.0}_{\pm 0.5}$ & $\uline{59.1}_{\pm 0.5}$ & $78.2_{\pm 1.2}$ & $\uline{41.4}_{\pm 0.1}$ & $\textbf{16.1}_{\pm 0.2}$ & $\uline{32.5}_{\pm 0.1}$ \\
        250K Frame & $\textbf{54.1}_{\pm 0.6}$ & $\textbf{56.1}_{\pm 0.3}$ & $\textbf{59.5}_{\pm 0.7}$ & $\textbf{77.1}_{\pm 0.7}$ & $\textbf{42.3}_{\pm 0.3}$ & $\textbf{16.1}_{\pm 0.1}$ & $\textbf{32.7}_{\pm 0.1}$ \\
        \bottomrule
   
    \end{tabular}
    
    \label{tab:dataset_scaling}
\end{table*}

\begin{tcolorbox}[colback=gray!10, colframe=black, title=RQ3 Summary:]
Our method shows a steady improvement in model performance as the data scale increases, demonstrating strong scalability and showing potential of utilizing massive training data from the combination of heterogeneous datasets.
\end{tcolorbox}

\section{Related works}
\label{section:related_work}

\paragraph{3D perception tasks}
3D perception focuses on analyzing and interpreting 3D data, including tasks such as detection~\cite{chen2017multi, yan2018second, sindagi2019mvx, qi2019deep, shi2019pointrcnn, vora2020pointpainting, huang2020epnet, shi2020pv, shi2020points, yoo20203d, yang20203dssd, cheng2021back, wang2021pointaugmenting, deng2021voxel, wang2021fcos3d, huang2021bevdet, bai2022transfusion, wang2022rbgnet, liang2022bevfusion, li2022deepfusion, yang2022boosting, liu2022petr, wang2022detr3d, li2022bevformer, wu2023virtual, wang2023mvcontrast, liu2023bevfusion, gao2023sparse, wang2023unitr}, segmentation~\cite{philion2020lift, zhu2021cylindrical, kim2022learning, wang2022cross, li2022bevformer, liu2023bevfusion, wang2023unitr}, tracking~\cite{yin2021center, wang2023tracking} and occupancy prediction~\cite{openoccupancy, huang2023tri, occmamba}. Detection involves identifying and localizing objects within a 3D scene, commonly applied in areas like autonomous driving and robotics. Segmentation refers to partitioning a scene into meaningful regions, using semantic segmentation to label each point with a class, or instance segmentation to separate objects of the same class, which is important in fields like medical imaging. Tracking, on the other hand, involves following the movement of objects over time, crucial in dynamic environments like surveillance and autonomous systems. Occupancy prediction, as a newly proposed task in recent years, involves predicting both voxel occupancy and semantic labels for each voxel, which is essential for generating complete models from partial data and supporting digital twin technology in industrial applications. 
\paragraph{Self-supervised pre-training for 3D vision}
Recently, amounts of works improve the performance of models for 3D Vision tasks via self-supervised pre-training on image and point cloud modalities. 
For unimodal methods, Contrastive learning~\cite{chen2020simple} and MAE~\cite{he2021maskedautoencodersscalablevision} are the most commonly used methods. 
Contrastive learning methods, which generate similar sample pairs to encourage similar samples to have have closer feature representations, are proven to be effective in image modalities~\cite{chen2020simple,henaff2020data,misra2020self,chen2020improved,he2019moco}. But for point cloud modality, ~\cite{xie2020pointcontrast,liu2020p4contrast,zhang2021self} are typically limited to object-level point clouds or smaller indoor scenes due to significant training challenges caused by large scale point clouds. 
On the other hand, MAE methods, which train the network to recover masked regions, helping the network capture feature information. Similar to works on the image modality~\cite{xie2022simmim,wei2022masked,he2021maskedautoencodersscalablevision}, some works~\cite{xie2022m} directly applying MAE to voxels, but other works~\cite{zhang2022point,yang2023gdmaegenerativedecodermae} modifies the network architecture for better information extraction.
For multi-modal methods, many works have used these two modalities in collaboration for pre-training. \cite{li2022simipu} used super-pixels as a unified encoding view, while ~\cite{chen2022bevdistill,sun2023calico,liu2023geomim} used BEV (Bird's Eye View), and~\cite{zhang2024hvdistill} combined both views for training. \cite{min2024multi} employed spatial grids as the view for comparing the two modalities. These works often focus on knowledge distillation from one modality to the other. However, \cite{zhou2023unidistilluniversalcrossmodalityknowledge} found that bidirectional collaborative distillation between the two modalities achieves the best results.
\paragraph{Multi-dataset training for vision tasks}
Combining multiple datasets for training is an effective method to scale up the amount of training data. However, naively merging different datasets can be harmful for model performance. Relevant studies have been proposed in both the 2D vision field and the point cloud 3D perception field. Researchers have used different normalization layers to differentiate between datasets~\cite{wang2022cross} or have performed category alignment~\cite{kim2022learning,zhou2022simple} to mitigate the impact of dataset gap. Recent studies~\cite{wu2024towards} have employed prompt techniques to further alleviate the negative impact caused by domain gaps between point clouds from different 3D datasets. 

\section{Conclusion and discussion}
\label{section:conclusion}
We propose a self-supervised pre-training framework with combined datasets for 3D perception in autonomous driving. Experimental results demonstrate the effectiveness of our methods in utilizing unlabeled data and mitigating the gap between datasets. The scaling experiments shows steady improvement on model performance as the training data volume increases, demonstrating strong scalability and the potential of enabling models to continually evolve toward foundation models with massive training data.

Due to limitations in public datasets and training resources, the scale of datasets and models in our research remains constrained. Our method offers insights for research institutions with access to extensive data and computational resources. For future works, we aim to further exploring the scaling effect of model size for 3D perception models. Additionally, we encourage research institutions to contribute more public datasets, so that the community can collaboratively build foundation models for the 3D perception field with our inspiration. 
{
    \small
    \bibliographystyle{ieeenat_fullname}
    \bibliography{main}
}
\clearpage
\appendix
\section{Hyperparameters \& configurations of pre-training}
For experiments in Table~\ref{tab:pretrain_effect} in section~\ref{section:exp_pretrain_effect}, we use a linear lr warm up and then use a cosine lr decay policy to reduce the lr to 0.01 times of the peak value. We conducted parameter search for learning rate and the final parameters are as in table~\ref{tab:pretrain_effect_param}. 
\begin{table}[htb]
    \centering
    \begin{tabular}{l|c}
         \toprule
         Parameter&Value\\
         \midrule
         Learning rate& 4e-4 \\
         Batch size& 32\\
         Training epochs& 50\\
         LR warm epoches&5\\
         Weight decay&0.01\\
         \bottomrule
         
    \end{tabular}
    \caption{Hyperparameters of pre-training in table~\ref{tab:pretrain_effect}}
    \label{tab:pretrain_effect_param}
\end{table}

For experiments in Table~\ref{tab:dataset_fusion}, ~\ref{tab:effect_prompt} and ~\ref{tab:dataset_scaling}, we keep the settings except for setting learning rate to 2e-4. The dataset volume and repeat sampling times in each epoch for experiments in Table~\ref{tab:dataset_scaling} are provided in Table~\ref{tab:dataset_sample}.

\begin{table*}[]
    \centering
    \begin{tabular}{l|cccccc}
         \toprule
         \multirow{2}*{Experiment}& \multicolumn{2}{c}{NuScenes}& \multicolumn{2}{c}{NuScenes}&\multicolumn{2}{c}{NuScenes}\\
        \cmidrule(r){2-3}\cmidrule(r){4-5} \cmidrule(r){6-7}       
        & Data volume&Repeat times& Data volume&Repeat times& Data volume&Repeat times\\
        \midrule
         40K Frame& 13,300&1&13,300&1&13,300&1 \\
         100K Frame& 28,130&1&22,680&1&50,000&1\\
         250K Frame& 28,130&4&22,680&4&200,000&1\\
         \bottomrule        
    \end{tabular}
    \caption{Configures of dataset sampling for pre-training in table~\ref{tab:dataset_scaling}.}
    \label{tab:dataset_sample}
\end{table*}
\section{Discussion on baselines \& related works}
In previous research, many pre-training methods for point cloud and image modality collaboration have been proposed. These methods transform images and point clouds into a unified perspective for joint training. Based on the chosen perspective, they can be categorized into three types: superpixel-based, BEV-based, and occupancy-based methods.  

\textbf{Occupancy-based} methods such as UniScene~\cite{min2024uniscenemulticameraunifiedpretraining} construct spatial occupancy grids from point clouds and reconstruct these grids using image modality data. These methods primarily focus on pre-training image networks, offering limited utility in multimodal 3D perception scenarios.  

\textbf{Superpixel-based} methods, such as SimIPU~\cite{li2022simipu}, divide images into superpixels and treat the corresponding point cloud regions within each superpixel as a single unit for comparison. However, these methods are incompatible with modern BEV-based 3D perception models.  

\textbf{BEV-based} methods unify image and point cloud information in the BEV perspective for comparison. The most representative works is BEVDistill~\cite{chen2022bevdistill}. GeoMIM~\cite{liu2023geomim} and UniDistill~\cite{zhou2023unidistilluniversalcrossmodalityknowledge} have modified this approach. Their training objectives often involves distilling 3D information from point clouds into the image modality or transferring knowledge from pre-trained image backbones to point cloud backbones. These methods typically use MSE loss to directly align the features of the student modality with those of the teacher modality. However, this setup is not suitable for training models in point cloud-image fusion scenarios from scratch. Additionally, these works rely on labeled data to train decoders for object-level distillation, which limits their scalability.  

The most rescent research, CALICO~\cite{sun2023calico}, compares superpixel-based and BEV-based methods, highlighting that BEVDistill-type approaches offer better performance and greater potential for improvement. CALICO uses image-point cloud contrastive loss and a point-cloud-only self-contrastive loss, emphasizing the training of point cloud networks. However, CALICO is not open-sourced and relies on relatively traditional backbone networks, making direct comparisons difficult.  

Given these reasons, we follow CALICO and adopt the most relevant BEVDistill method as our experimental baseline, modifying it to fit a self-supervised learning framework by removing the object-level distillation loss that depends on labeled data.
\section{Details of downstream tasks}
\subsection{3D object detection}

In this section, we describe the detailed experimental setups used in our downstream tasks. To leverage the advantages of multimodal pretraining, we employed the BEVFusion multimodal framework to perform downstream 3D object detection and BEV map segmentation tasks. For the image and LiDAR backbones, we utilized Swin-Transformer and DSVT, with an image size of [256, 704] and a voxel size of [0.3, 0.3, 8], respectively. The resulting BEV feature maps for LiDAR and Camera are sized [180, 180, 128] and [180, 180, 80], respectively.
We concatenated these two BEV feature maps and applied a fusion layer, followed by a BEV backbone and an FPN network. 
For 3D detection, we used the Transfusion head with $8\times 3$ batches on the NuScenes dataset and the DSVT head with $8\times 4$ batches on the Waymo dataset. For both datasets, we employed the AdamW optimizer with a cyclic scheduler and a starting learning rate of $1 \times10^{-4}$.
Additionally, we used ground-truth copy-paste data augmentation during training, which was disabled in the last five epochs.

When we get pre-trained checkpoint, we just load parameters of Swin-Transformer and DSVT backbones. For "Correspond" setting, we fix the prompt to “NuScenes”, and "Once" for "Wrong" setting. For "Random" setting, we train the prompt from scratch.

\subsection{BEV map segmentation}
For BEV map segmentation, we employ the same framework as 3D object detection, but with a different head. We use the same BEVSegmentationHead as BEVFusion, first obtaining $200\times 200$ BEV features through linear interpolation, and then apply a convolutional layer to predict the class of each grid. We employed the AdamW optimizer with a cyclic scheduler, starting with a learning rate of $1 \times 10^{-4}$. The model's performance was evaluated on the NuScenes dataset.  Finally, the handling of the pre-trained checkpoint follows the same approach as in 3D object detection.

\subsection{3D object tracking}
In the task of 3D object tracking, we utilize SimpleTrack~\cite{pang2022simpletrack}, a "tracking-by-detection" framework directly leveraging detection results as input. Initially, the input detections are pre-processed to select the ones to be used for tracking. Subsequently, a motion model is employed to predict and update the states of the tracked objects. After generating predictions across frames, a data association step is performed to link detections with existing tracklets. In addition to the core algorithmic steps, "birth", "death" and "output" policies are implemented to manage the life-cycle of detections and tracklets. For the nuScenes dataset, we adhere to the default hyper-parameter settings in SimpleTrack, which include using 3D GIoU as the association metric and applying a score threshold of 0.01 for output bounding boxes. These settings are well-suited for detectors incorporating LiDAR modalities\cite{pang2023standing}.

\subsection{Occupancy prediction}
In Occupancy Prediction task, we utilize the OpenOccupancy dataset~\cite{openoccupancy}, which is an occupancy annotation dataset derived from the nuScenes dataset and adheres to its data format. The dataset includes 700 training sequences and 150 validation sequences, with annotations provided for 17 different classes. For each input frame, we use six surround-view camera images resized to [256, 704] as visual input and fuse ten frames of LiDAR points covering a spatial range of [-51.2m, 51.2m] along both the X and Y axes and [-2.0m, 6.0m] along the Z axis. The occupancy annotations are mapped onto a voxel grid of dimensions $512\times512\times40$, with each voxel measuring 0.2 meters. In terms of model architecture, we build upon the OccMamba~\cite{occmamba} framework, setting mamba features dimension to 128 and replacing the backbones for the image and LiDAR inputs with Swin-Transformer and DSVT, respectively, as referenced in the aforementioned works. For our training strategy, we increase the learning rate to $5\mathrm{e}{-4}$ based on OccMamba. Specifically, during training with the dataset prompt, we fix the learnable prompt in PromptNorm to ensure better results on this challenging task. All other model and training configurations remain consistent with the original OccMamba settings.
\section{Analysis of prompts}
\begin{figure*}
    \centering
    \includegraphics[width=0.99\linewidth]{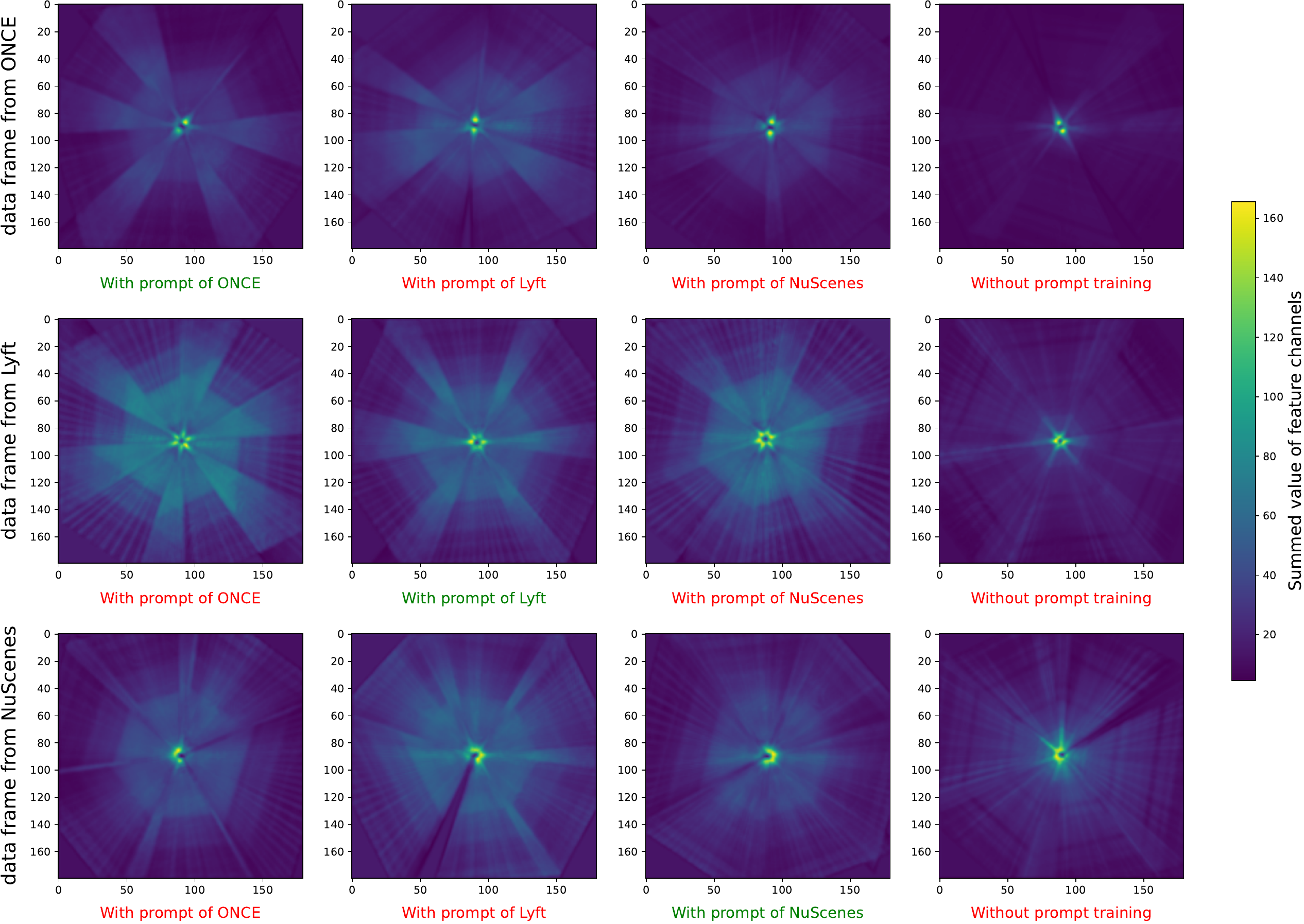}
    \caption{BEV heatmaps of image modality when testing the models by applying various prompts or using the model without prompt training.}
    \label{fig:prompt_effect_full}
\end{figure*}

To further investigate the role of prompts in the model, we visualized the image BEV heatmap, as shown in Figure 4~\ref{fig:prompt_effect_full}, where the x-axis represents the forward direction of the vehicle. We extracted one sample each from the NuScenes, ONCE, and Lyft datasets and tested them using networks with different prompts or without prompts. The visualization reveals that the image BEV features tend to cluster in specific regions across all three datasets.

In these datasets, the image data are collected from six camera views: front, front-left, front-right, rear-left, rear-right, and rear. Due to calibration issues, slight tilts may occur, and variations in viewing angles and other camera parameters exist among the datasets. Figure 4~\ref{fig:prompt_effect_full} highlights that when the correct prompt is applied, the network can effectively focus feature information on the central regions captured by the cameras, thereby extracting more valuable features. In contrast, using incorrect prompts still produces a clustering effect, but the focus is less precise. Without employing the prompt training strategy during training, the feature distribution becomes highly disorganized or the model fails to capture meaningful features.  

This visualization underscores the importance of prompts in guiding the model to align features accurately with the input data's intrinsic structure address the domain gap between datasets.

\end{document}